# A Context-Aware Middleware for Medical Image Based Reports

## An approach based on image feature extraction and association rules

Érick O. Rodrigues, José Viterbo and Aura Conci
Department of Computer Science
Universidade Federal Fluminense
Niterói, Rio de Janeiro
erickr@id.uff.br

Trueman MacHenry
Departament of Mathematics & Statistics,
York University
Keele St 4700, Toronto, ON M3J 1P3, Canada
machenry@mathstat.yorku.ca

**Abstract—This work proposes a context-aware middleware for medical workflow organization and efficiency improvement. In hospitals, laboratories and teleradiology companies, each physician or technician is specialized in a specific kind of diagnosis or analysis. Therefore, certain types of medical images are often forwarded to a certain physician or a certain group. This forwarding is time consuming. That is, repeatedly deciding who would be the best physician, whether he is available at a certain moment given a certain context is exhaustive and may be very inefficient. Thus, the proposed middleware has the ability to process and collect data from images analyzed by each medical staff. Based on the collected data and current clinical context, the middleware is able to infer who would be the best fit staff to receive a certain incoming medical image.**

## I. Introduction

In everyday clinical practice, physicians and nurses spend great part of their working time on inefficient communications. We define as inefficient communications any content or message that has to be passed on among the medical staff in a daily basis regarding distinct clinical contexts and, in this case, that are related to the overall medical care. That is, for instance, passing data of a certain patient to a physician or a nurse, such as information of which nurse is available at the moment for a clinical action such as a surgery, etc.

Besides these type of common communications there is also a specific kind, which is the medical image forwarding. In hospitals, laboratories and teleradiology companies, some physicians or technicians are specialists in analyzing, processing or producing reports from a certain type of image. For instance, even when regarding just a certain modality such as Computed Tomography (CT), some physicians are better at giving a certain diagnosis to a certain type of cancer. Furthermore, a group of physicians may be able to manually segment certain organs with better precision than the remaining. A major issue arises with these particularities. The time consumption related to analyzing the medical context, checking whether someone is available and deciding which group or physician is the most fit to process a certain medical image is huge, besides being a financially costly procedure.

It would be certainly more efficient to automate these repetitive processes such as the medical image forwarding. Therefore, this work proposes a novel middleware capable of (1) collecting important information from users (physicians, technicians, etc) based on features extracted from images processed by them, (2) inferring and creating association rules from the extracted data and (3) autonomously forwarding images to the proper staff while regarding the clinical context, which is very complex.

## II. Fundamentals

### A. Context-Aware Middlewares

Context-aware applications have been often defined and discussed in various works such as in Nihei [1] and Baldauf et al. [2]. Context information can be defined as any additional information that can be used to improve the behavior of a service in an arbitrary situation. That is, if such additional information is disregarded, the service would still operate. However, when regarding context information, the service would operate in a more efficient or appropriate manner [3]. Therefore, a service or application that incorporates any context information can be defined as context-aware.

Context information is often but not strictly obtained from sensors. For instance, a mobile device is a rich source of sensors, usually containing a camera, gyroscope, network connection and GPS [4]. Besides, it is usually carried by their owners in a daily basis, which turns them into perfect data collectors. All the data retrieved from these devices can be used to infer the desired context in an application. However, sensors are not the only source of context information. Datasets containing information from staffs of an institution [5], data collected from social networks and overall applications [4, 6], user driven interfaces for data collection [5] are also regarded as major context information providers.

### B. Medical Images and Reports

In the context of a hospital, several physicians and technicians work concurrently to provide suitable reports for incoming medical images. However, it is evident that one or a certain group of physicians produces better reports or are specialists in a very specific type of content within the images as previously discussed. That is, if we regard X-ray breast images that contain a specific type of tumor, one physician, for instance, may excel the others on the classification of that tumor. Thus, a middleware that automatically evaluates the current context of the hospital (e.g., best available physicians related to the medical image, where they are and whether they are available) and sends the desired image to the right physician at the right time would certainly increase the working efficiency of the medical staff and institution in general. Besides, this type of middleware can also be employed in medical laboratories, teleradiology companies, etc.

In a real life scenario, what often happens in very specific scenarios such as the one previously addressed is that the medical staff waits for the most suited physician in order to finish the report. The time spent on intercommunication to complete that objective is huge. As a premise of the ubiquitous computing, the proposed middleware would solve that issue and should be imperceptible from the point of view of the medical staff, which are the sole users in this case [7].

Each modality of medical images are significantly different from each other. Therefore, methodologies within the visual computing field are usually developed for a single modality, such as the ones in [7-12]. However, the proposed middleware needs to be general, that is, there is no a priori knowledge as to what to search in the images, neither what kind of images they are. Thus, the proposed methodology is based on robust image features and mined data. The application should recognize the behavior of each physician or technician related to their evaluated images in order to infer further actions such as image forwarding.

## III. Related Works

Several works have already proposed middlewares for elderly homecare or assisted living [13-15]. Other authors have already proposed middlewares for wearables in a healthcare perspective [16,17]. Besides, the work of Cheng et al. [5] proposed a medical context-aware instant messaging architecture to be employed in hospitals aiming to reduce the time spent on intercommunication.

This work is based on the idea proposed by Cheng et al. [5]. In their work, the authors proposed a middleware for the intercommunication in hospitals. They state that between 44000 to 98000 Americans die each year as the result of medical errors. Furthermore, communication or also the lack of it has been shown to be a major contributor to these medical errors. Moreover, Moss et al. [18] conducted a survey regarding nurse communication patterns at four operating rooms suites in three hospitals, showing that the most frequent mode of communication is face to face, followed by telephone and intercom. The authors did also point out that physicians spend 33% of their working time in communications while nurses spend 22%. Thus, in order to overcome this issue, Cheng et al. proposed an instant messaging application that is aware of the hospital context.

The proposed messaging application uses ontology, medical and network related information as context information. The medical database, for instance, contains data of the medical staff and information such as work schedule as well as if each member is available at the requested moment. The authors built their own traditional instant messaging (IM) application, which contain a server and clients that connect directly to the server, further extending it with context-aware modules. The context-aware part of the software is associated to the transport module, which sends and receive messages, containing options for setting pre-defined rules. Besides, along with the pre-defined rules, the module works with the context ontology, which covers properties of the entities (medical staff) and how they relate to each other.

In this work, we propose a slightly similar architecture if compared to the work of Cheng et al. [5]. However, the main difference is that we propose a methodology for predicting the path that an arbitrary medical image has to follow, instead of simple text messages.

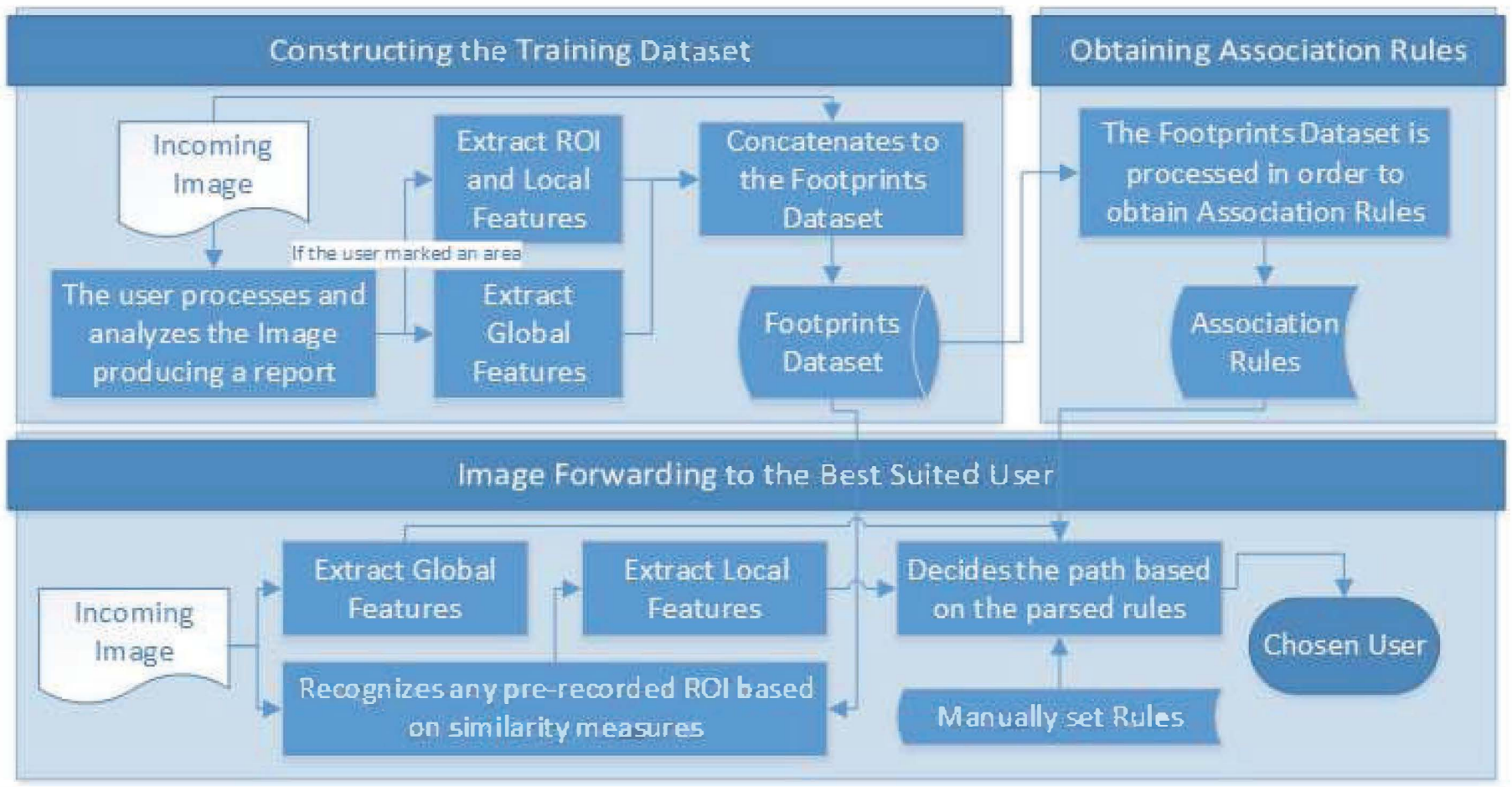

Fig. 1. Overview of the proposed methodology.

## IV. METHODOLOGY

The proposed approach comprises topics from the fields of visual computing, data mining and ubiquitous computing. At first, the methodology should learn from the patterns produced by every related medical staff. That is, every physician has two main forms of interacting with an image waiting to be processed: (1) they can just visualize the image and produce a medical report or (2) they can make markings within the image and produce a medical report. The manual markings separate groups of pixels of the image, which in this specific case can also be called regions of interest (ROI).

Thereafter, some information regarding the behavior of the physicians or technicians are recorded in the application along with features extracted from the images analyzed by them. The recorded data are, namely: (1) the identifier of the physician, (2) the analyzed image along with its ROI and (3) features extracted from both the entire image and ROI. We define the features extracted from the image as global features while the ones extracted from the manual markings are defined as local features. As global features we have selected: (1) a coefficient of smooth variation [19], (2) an arithmetic mean [19], (3) moments of the co-occurrence matrix [19], (4) geometric moments [19] and (5) moments of the co-occurrence matrix of a gradient of the image obtained from the Sobel edge detection [20]. As local features we have selected all the features used in the global case along with the (1) run percentage and (2) grey level non-uniformity [19]. These two latter features are more computationally expensive than the remaining.

After the data extraction, these are considered footprints of the user (physician or technician). Thus, association rules with relation to the identifier of the various users are mined from all the recorded footprints. These rules are further applied to predict and direct the path of an incoming image to a specific user. Eventually, the extracted ROIs are joined to form an atlas or template that is used to search for the same area in an incoming image. That is, if a physician is used to segment a certain type of tumor then the collection of the already segmented tumors are joined (arithmetic mean) to form an image that is further used to search for similar spots on incoming images. This information is also used to bias the path of the image. The entire methodology is summarized in Figure 1.

As shown in Figure 1, the methodology can be divided in three main steps: (1) construction of the training data, (2) obtainment of the association rules and (3) the actual image forwarding. These three steps can be executed fully separated as long as the step 2 is executed after 1 and 3 after 2. Moreover, the module called "Manually set Rules" stands for the appliance of rules that can be directly added by the user, preferably based on some ontology context. For instance, in a hospital, information regarding the appropriate department can be used to filter possible paths for the image.

### A. No Response Tractability

A module of the architecture is responsible for listening and waiting for a response. That is, once the image has been forwarded to the best fit user at the first iteration, the sender waits for a response of the user, as shown in Figure 2 (start button). The response can be (1) positive, when the user has produced or will produce a medical report based on the image or (2) negative when the user did not or will not produce a medical report. A timer is also set for stipulating a maximum waiting time in cases where no response has been sent from the user. If the response is negative or no response has been sent, the algorithm recalculates a new path for the image and repeats the same listening and waiting process.

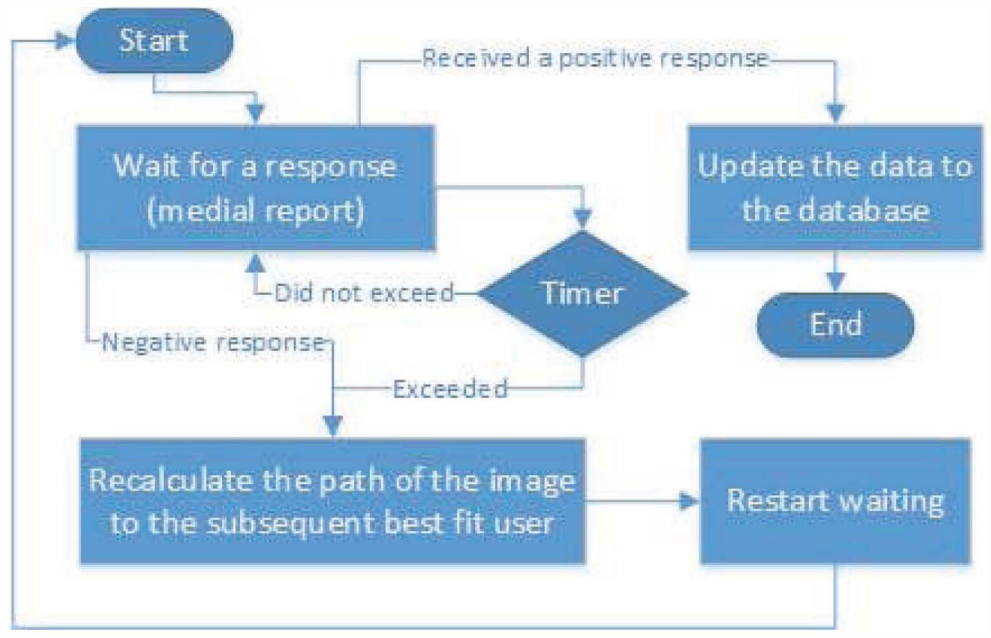

Fig. 2. Response waiting module.

In a worst case scenario, an image is sent to a user that does not respond. In this case, the image is not analyzed at least until the timer is exceeded. To avoid situations like that, the image is actually sent to a certain number of users and they are asked to send a positive response, that is, an agreement that they will be responsible for the current forwarded image. The user that sends a positive reply and is best fit to process a report based on the current image will proceed with the report production.

The selected features to be extracted from the evaluated images are described in the following subsections. We have chosen features that altogether are able to distinguish differences not only between modalities but between specific changes in parts of the images.

### B. ROI Recognition

Every time a subset of pixels of an image is marked by the user, this subset (or ROI) is compared using similarity measures to all the ROIs previously selected by the same user. If the comparison reaches a certain threshold, then the compared ROIs are considered to be of the same kind. Thus, these two or more ROIs are joined together to a single image by an arithmetic mean. In a further comparison, the selected ROI is compared to the mean image instead. Figure 3 illustrates these processes.

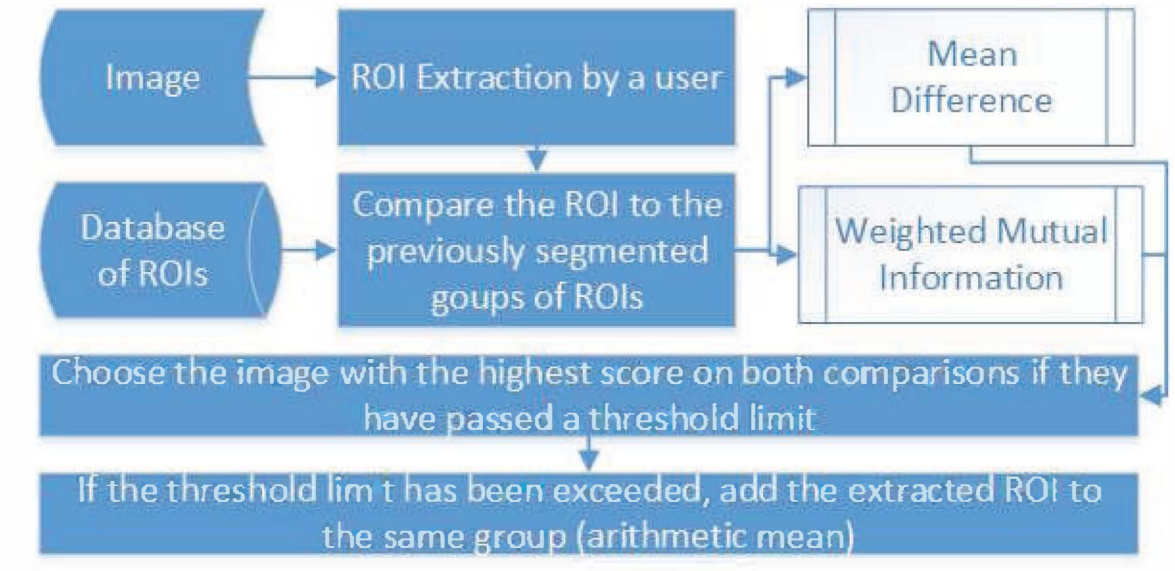

Fig. 3. ROI recognition and association.

The proposed methodology compares the ROI images using two similarity measures: Mean Difference (MD) and Weighted Mutual Information (WMI) [19]. Thus, the highest score on both measures dictates the group that the image belongs to, as long as if the measure has reached an established threshold. This threshold is variable according to the type of regarded images or modalities, and should be empirically chosen. The best result is given by the minimization of the chosen equation (MD or WMI). When the images are the same, the mean difference similarity measure returns 0 [19].

## V. Further Experiments

The proposed middleware can be further evaluated in a real scenario if applied to hospitals, laboratories or teleradiology companies. For instance, the time spent to produce a report of specific medical images with the proposed middleware can be compared to the time spent to complete the same task without it. Besides, the overall report production can be compared in both occasions, when employing and not employing the middleware. We are greatly confident that the proposed middleware would speed up not only the analysis of specific cases but the production as a whole, since trivial intercommunication is reduced.

Furthermore, the parameters of the extracted features can also be evaluated from the data of a real application, as well as the extracted features themselves. Certain features and parameters would perform better in a certain hospital, for instance. Therefore, if the middleware is applied to distinct organizations, then robustness of the proposed features can also be analyzed from a general view. Reducing the amount of extracted features would reduce the current processing time of the approach, which is next to 7 minutes for each processed image in a simple personal computer.

Moreover, the middleware can also serve as a historic of actions, saving more specific traces of several physicians and technicians (such as their action and what they are doing within images). Perhaps this could lead to several future studies with regard to the staff that does their work better than the remaining, in a sense to study their behavior aiming to potentially reproduce their methodology. Furthermore, this historical data can also be used for auditing, as well as in data mining, in several different ways. Unfortunately, our middleware is still in development and could not be currently applied in a real case scenario.

## VI. Conclusion

The proposed middleware has the potentiality of greatly reducing the time consumption of daily medical intercommunications and, consequently, improving the overall efficiency of medical care. However, due to its complexity, the middleware is still under development and experimental data are also to be collected in the near future.

We are the first to propose a general and functional middleware for medical images forwarding. Some of the addressed features work exceptionally well with medical images but are computationally expensive. As a future work we would like to analyze the feature selection more properly, evaluating the positive contribution to the approach versus the processing time consumed by each one.